\documentclass[11pt,a4paper]{article}
\usepackage[nohyperref]{acl2017}
\usepackage{times}
\usepackage{latexsym}

\usepackage{url}

\usepackage{amsmath}
\usepackage{multirow}
\usepackage{url}
\usepackage{graphicx}
\usepackage[ruled,vlined]{algorithm2e}
\usepackage{amssymb}
\usepackage{amsfonts}
\usepackage{todonotes}
\usepackage{fancyvrb}
\usepackage{tikz}
\aclfinalcopy 

\title{A Mixture Model for Learning Multi-Sense Word Embeddings}

\author{
Dai Quoc Nguyen${}^{1}$, Dat Quoc Nguyen${}^{2}$, Ashutosh Modi${}^{1}$, Stefan Thater${}^{1}$, Manfred Pinkal${}^{1}$\\
\\
${}^{1}$Department of Computational Linguistics, Saarland University, Germany\\
{\tt{{\{daiquocn, ashutosh, stth, pinkal\}@coli.uni-saarland.de}}} \\
${}^{2 }$Department of Computing, Macquarie University, Australia \\
{\tt{{dat.nguyen@students.mq.edu.au}}}
}

\begin{document}
\maketitle

\begin{abstract}

Word embeddings are now a standard technique for inducing meaning representations for words. For getting good representations, it is important to take into account different senses of a word. In this paper, we propose a mixture model for learning multi-sense word embeddings. Our model generalizes the previous works in that it allows to induce different weights of different senses of a word. The experimental results show that our model outperforms previous models on standard evaluation tasks.
 
\end{abstract}

\section{Introduction}
Word embeddings have shown to be useful in various NLP tasks such as sentiment analysis, topic models, script learning, machine translation, sequence labeling and parsing \cite{socher-EtAl:2013,Sutskever:2014,modititov:2014,nguyen2015,NguyenALTA2015,modi:2016,ma-hovy:2016:P16-1,jPTDP2017,ashutoshTacl2017}.
A word embedding captures the syntactic and semantic properties of a word by representing the word in a form of a real-valued vector \cite{MikolovCoRR2013,MikolovSCCD13nips,Pennington2014,Levy:2014}. 

However, usually word embedding models do not take into account lexical ambiguity. 
For example, the word $bank$ is usually represented by a single vector representation for all senses including \textit{sloping land} and \textit{financial institution}.
Recently, approaches have been proposed to learn multi-sense word embeddings, where each sense of a word corresponds to a sense-specific embedding.
\newcite{Reisinger:2010}, \newcite{Huang:2012} and \newcite{Wu:2015} proposed methods to cluster the contexts of each word and then using cluster centroids as vector representations for word senses. 
\newcite{neelakantan:2014}, \newcite{tian-EtAl:2014}, \newcite{li-jurafsky:2015} and \newcite{chen:2015} extended  Word2Vec models \cite{MikolovCoRR2013,MikolovSCCD13nips} to learn a vector representation for each sense of a word. 
\newcite{chen-liu-sun:2014}, \newcite{iacobacci:2015} and \newcite{flekova-gurevych:2016:P16-1} performed word sense induction using external resources (e.g., WordNet, BabelNet) and then learned sense embeddings using the Word2Vec models. 
\newcite{RotheS15} and \newcite{PilehvarC16} presented methods using pre-trained word embeddings to learn embeddings from WordNet synsets.
\newcite{Cheng:2015}, \newcite{Liu:2015}, \newcite{LiuPengfei:2015} and \newcite{Zhang2016} directly opt the Word2Vec Skip-gram model \cite{MikolovSCCD13nips} for learning the embeddings of words and topics on a topic-assigned corpus.

One issue in these previous works is that they assign the same weight to every sense of a word.
The central assumption of our work is that each sense of a word given a context, should correspond to a mixture of weights reflecting  different association degrees of the word  with multiple senses in the context.
The mixture weights will help to model word meaning better.

In this paper, we propose a new model for learning \textbf{M}ulti-\textbf{S}ense \textbf{W}ord \textbf{E}mbeddings (\textsc{mswe}). 
Our \textsc{mswe} model learns vector representations of a word based on a mixture of its sense representations.  
The key difference between \textsc{mswe} and other models is that we induce the weights of senses while jointly learning the word and sense embeddings.
Specifically, we train a topic model \citep{Blei2003} to obtain the topic-to-word and document-to-topic probability distributions which are then used to infer the weights of topics. We use these weights to define a compositional vector representation for each target word to predict its context words. \textsc{mswe} thus is different from the topic-based models \citep{Cheng:2015,Liu:2015,LiuPengfei:2015,Zhang2016}, in which we do not use the topic assignments when jointly learning vector representations of words and topics. Here we not only learn vectors based on the most suitable topic of a word given its context, but  we also take into consideration all possible meanings of the word.   

The main contributions of our study are:
(i) We introduce a mixture model  for learning word and sense embeddings (\textsc{mswe}) by inducing mixture weights of word senses.
(ii) We show that \textsc{mswe} performs better than the baseline Word2Vec Skip-gram 
and other embedding models on the word analogy task \cite{MikolovCoRR2013} and the word similarity task \cite{Reisinger:2010}.

\section{The mixture model}
\label{sec:mswe}

In this section, we present the mixture model for learning multi-sense word embeddings. 
Here we treat topics as senses. The model learns a representation for each word using a mixture of its topical representations.

Given a number of topics and a corpus $D$ of documents $d = \{w_{d,1}, w_{d,2},..., w_{d,M_d}\}$, we apply a topic model \cite{Blei2003} to obtain the topic-to-word $\Pr(w|t)$ and document-to-topic $\Pr(t|d)$ probability distributions. We then infer a weight for the $m^{th}$ word $w_{d,m}$ with topic $t$ in document $d$:
\begin{equation}
\lambda_{d,m,t} = \Pr(w_{d,m}|t) \times \Pr(t|d)
\label{equal:weight}
\end{equation}

We define two \textsc{mswe} variants: \textsc{mswe-1} learns vectors for words based on the most suitable topic given  document $d$  while \textsc{mswe-2} marginalizes over all senses of a word to take into account all possible senses of the word:
\begin{eqnarray}
\textsc{mswe-1:} & \boldsymbol{s}_{w_{d,m}} &= \dfrac{\boldsymbol{v}_{w_{d,m}} + \lambda_{d,m,t'} \times \boldsymbol{v}_{t'}}{1 + \lambda_{d,m,t'}} \nonumber \\
\textsc{mswe-2:} & \boldsymbol{s}_{w_{d,m}} &= \dfrac{\boldsymbol{v}_{w_{d,m}} + \sum_{t=1}^{T} \lambda_{d,m,t} \times \boldsymbol{v}_{t}}{1 + \sum_{t=1}^{T} \lambda_{d,m,t}} \nonumber
\end{eqnarray}
where $\boldsymbol{s}_{w_{d,m}}$ is the compositional vector representation of the $m^{th}$ word $w_{d,m}$ and the topics in document $d$; $\boldsymbol{v}_w$ is the target vector representation of a word type $w$ in vocabulary $V$; $\boldsymbol{v}_{t}$ is the vector representation of topic $t$; $T$ is the number of topics; $\lambda_{d,m,t}$ is defined as in Equation \ref{equal:weight}, and in \textsc{mswe-1} we define $t' = \underset{t}{\arg\max} \ \lambda_{d,m,t}$.  

We learn representations by minimizing the following negative log-likelihood function:
\begin{equation}
\resizebox{.89\hsize}{!}{$
\mathcal{L}  = - \sum\limits_{d \in D} \sum\limits_{m=1}^{M_d} \sum\limits_{\substack{-k \leq j \leq k \\ j \neq 0}} \log \Pr(\tilde{\boldsymbol{v}}_{w_{d,m+j}}|\boldsymbol{s}_{w_{d,m}})
$}
\label{equal:1}
\end{equation}

\noindent where the $m^{th}$ word $w_{d,m}$ in document $d$ is a target word while the $(m+j)^{th}$ word $w_{d,m+j}$ in document  $d$  is a context word of $w_{d,m}$ and $k$ is the context size. In addition, $\tilde{\boldsymbol{v}}_{w}$ is the context vector representation of the word type $w$. 
The probability $\Pr(\tilde{\boldsymbol{v}}_{w_{d,m+j}}|\boldsymbol{s}_{w_{d,m}})$ is defined using the softmax function as follows:
\begin{equation*}
\Pr(\tilde{\boldsymbol{v}}_{w_{d,m+j}}|\boldsymbol{s}_{w_{d,m}}) = \frac{\exp(\tilde{\boldsymbol{v}}_{w_{d,m+j}}^\mathsf{T} \boldsymbol{s}_{w_{d,m}})}{\sum_{c' \in V} \exp(\tilde{\boldsymbol{v}}_{c'}^\mathsf{T} \boldsymbol{s}_{w_{d,m}})} 
\label{equal:2}
\end{equation*}

Since computing $\log\Pr(\tilde{\boldsymbol{v}}_{w_{d,m+j}}|\boldsymbol{s}_{w_{d,m}})$ is expensive for each training instance, we approximate $\log\Pr(\tilde{\boldsymbol{v}}_{w_{d,m+j}}|\boldsymbol{s}_{w_{d,m}})$ in Equation \ref{equal:1} with the following negative-sampling objective \cite{MikolovSCCD13nips}:
\begin{eqnarray}
\mathcal{O}_{d,m,m+j}& = &\log \sigma \left(\tilde{\boldsymbol{v}}_{w_{d,m+j}}^\mathsf{T} \boldsymbol{s}_{w_{d,m}}\right) \nonumber \\ &+& \sum_{i=1}^{K} \log \sigma \left(-\tilde{\boldsymbol{v}}_{c_i}^\mathsf{T} \boldsymbol{s}_{w_{d,m}}\right)
\label{equal:3}
\end{eqnarray} 

\noindent where 
each word $c_i$ is sampled from a noise distribution.\footnote{We use an unigram distribution raised to the 3/4 power \cite{MikolovSCCD13nips} as the noise distribution.}
In fact, \textsc{mswe} can be viewed as a generalization of the well-known  Word2Vec Skip-gram model with negative sampling \cite{MikolovSCCD13nips} where all the mixture weights $\lambda_{d,m,t}$ are set to zero. The models are trained using Stochastic Gradient Descent (SGD).


\section{Experiments}
\label{sec:exp}

We evaluate \textsc{mswe} on two different tasks: word similarity and word analogy. We also provide experimental results obtained by the baseline Word2Vec Skip-gram model and other previous works. 

Note that not all previous results are mentioned in this paper for comparison because the training corpora used in most previous research work are much larger than ours \citep{baroni:2014,li-jurafsky:2015,schwartz:2015,Levy:2015}. Also there are differences in the pre-processing steps that could affect the results. 
We could also improve obtained results by using a larger training corpus, but this is not central point of our paper. The objective of our paper is that the embeddings of topic and word can be combined into a single mixture model, leading to good improvements as established empirically.

\subsection{Experimental Setup}

Following \citet{Huang:2012} and \citet{neelakantan:2014}, we use the  Wesbury Lab Wikipedia corpus \citep{shaoul2010westbury} containing over 2M articles with about 990M  words for training.
In the preprocessing step, texts are lowercased and tokenized, numbers are mapped to 0, and punctuation marks are removed.
We extract a vocabulary of  200,000 most frequent word tokens  from the pre-processed corpus.
Words not occurring in the vocabulary are mapped to a special token \textsc{unk}, in which we use the embedding of \textsc{unk} for unknown words in the benchmark datasets.

We firstly use a small subset extracted from the \textsc{ws353} dataset \citep{Finkelstein:2001} to tune the hyper-parameters of the baseline Word2Vec Skip-gram model  for the word similarity task (see Section \ref{wordsim} for the task definition). We then directly use the tuned hyper-parameters for our \textsc{mswe} variants.
Vector size is also a hyper-parameter. While some approaches use a higher number of dimensions to obtain better results, we fix the vector size to be 300 as used by the baseline for a fair comparison.
The vanilla Latent Dirichlet Allocation (LDA) topic model \citep{Blei2003} is not scalable to a very large corpus, so we explore faster online topic models developed for large corpora. 
We train the online LDA topic  model \citep{Hoffman:2010} on the training corpus, and use the output of this topic model to compute the mixture weights as in Equation \ref{equal:weight}.\footnote{We use default parameters in \textit{gensim} \protect{\citep{rehurek_lrec}} for the online LDA model.} We also use the same \textsc{ws353} subset to tune the numbers of topics $T \in \{50, 100, 200, 300, 400\}$. We find that the most suitable numbers are $T = 50$ and $T = 200$ then used for all our experiments.
Here we learn 300-dimensional embeddings with the fixed context size $k=5$ (in Equation \ref{equal:1}) and $K = 10$ (in Equation \ref{equal:3}) as used by the baseline. 
During training, we randomly initialize model parameters (i.e. word and topic embeddings) and then learn them by using SGD with the initial learning rate of 0.01.

\subsection{Word Similarity}\label{wordsim}

The word similarity task evaluates the quality of word embedding models \cite{Reisinger:2010}. For a given dataset of word pairs, the evaluation is done by calculating correlation between the similarity scores of corresponding word embedding pairs with the human judgment scores. Higher Spearman's rank correlation ($\rho$) reflects better word embedding model. 
We evaluate \textsc{mswe} on standard datasets (as given in Table \ref{tab:datasets}) for the word similarity evaluation task.

\setlength{\abovecaptionskip}{3pt plus 1pt minus 1pt} 
\begin{table}[!t]
\centering
\setlength{\tabcolsep}{0.25em}
\resizebox{7.5cm}{!}{
\begin{tabular}{l|c|l}
\hline
Dataset & Word pairs & Reference \\
\hline
\textsc{ws353} & 353 & \newcite{Finkelstein:2001} \\
\textsc{SimLex} & 999 &  \newcite{Hill:2015} \\
\textsc{scws} & 2003 & \newcite{Huang:2012} \\
\textsc{rw} & 2034 & \newcite{luong:2013}\\
\textsc{men} & 3000 & \newcite{Bruni:2014} \\
\hline
\end{tabular}
}
\caption{The benchmark datasets. \textsc{ws353}: WordSimilarity-353. \textsc{rw}: Rare-Words. \textsc{SimLex}: SimLex-999. \textsc{scws}:  Stanford's Contextual Word Similarities. \textsc{men}: The MEN Test Collection. Each dataset contains similarity scores of human judgments for pairs of words.}
\label{tab:datasets}
\end{table}

Following \citet{Reisinger:2010}, \citet{Huang:2012}, \citet{neelakantan:2014}, we compute the similarity scores for a pair of words $\left(w, w'\right)$ with or without their respective contexts $\left(c, c'\right)$ as:

\vspace{-5pt}

{\small
\begin{eqnarray}
&&GlobalSim\left(w, w'\right) = \cos\left(\boldsymbol{v}_{w}, \boldsymbol{v}_{w'}\right)  \nonumber\\
&&AvgSim\left(w, w'\right) = \frac{1}{T^2} \sum_{t=1}^{T} \sum_{t'=1}^{T} \cos\left(\boldsymbol{v}_{w,t}, \boldsymbol{v}_{w',t'}\right) \nonumber\\
&& AvgSimC\left(w, w'\right) \nonumber \\ &&= \frac{1}{T^2} \sum_{t=1}^{T} \sum_{t'=1}^{T} \Big(\delta\left(\boldsymbol{v}_{w,t}, \boldsymbol{v}_{c}\right) \times \delta\left(\boldsymbol{v}_{w',t'}, \boldsymbol{v}_{c'}\right)  \nonumber \\
&&  \ \ \ \ \ \ \ \ \ \ \ \ \ \ \ \ \ \ \ \ \ \ \ \ \ \  \times \cos\left(\boldsymbol{v}_{w,t}, \boldsymbol{v}_{w',t'}\right)\Big) \nonumber
\end{eqnarray}
}

\vspace{-5pt}

\noindent where $\boldsymbol{v}_{w}$ is the vector representation of the word $w$, 
$\boldsymbol{v}_{w,t}$ is the multiple representation of the word $w$ and the topic $t$,
$\boldsymbol{v}_c$ is the vector representation of the context $c$ of the word $w$.
And $\cos\left(\boldsymbol{v}, \boldsymbol{v'}\right)$ is the cosine similarity between two vectors $\boldsymbol{v}$ and $\boldsymbol{v'}$. 
For our experiments, we set $\boldsymbol{v}_{w,t} = \boldsymbol{v}_{w} \oplus \left(\Pr(w|t) \times \boldsymbol{v}_{t}\right)$ and 
 $\boldsymbol{v}_c = \left(\frac{1}{|c|} \sum_{w \in c} \boldsymbol{v}_w\right) \oplus \left(\sum_{t} \Pr\left(t|c\right) \times \boldsymbol{v}_t\right)$, in which $\oplus$ is the concatenation operation and 
$\Pr\left(t|c\right)$ is inferred from the topic models by considering context $c$ as a document.
$GlobalSim$ only regards word embeddings, while $AvgSim$ considers multiple representations to capture different meanings (i.e. topics) and usages of a word.
$AvgSimC$ generalizes $AvgSim$ by taking into account the likelihood $\delta\left(\boldsymbol{v}_{w,t}, \boldsymbol{v}_{c}\right)$ that word $w$ takes topic $t$ given context $c$. $\delta\left(\boldsymbol{v}, \boldsymbol{v'}\right)$ is the inverse of the cosine distance from $\boldsymbol{v}$ to $\boldsymbol{v'}$ \cite{Huang:2012,neelakantan:2014}.  
{
\begin{table}[!t]
\centering
\setlength{\tabcolsep}{0.25em}
\resizebox{7.8cm}{!}{
\begin{tabular}{l|l|l|l|l|l}
\hline
Model & \textsc{rw} & \textsc{SimLex} & \textsc{scws} & \textsc{ws353} & \textsc{men} \\
\hline
\hline
\newcite{Huang:2012} & -- & -- & 58.6 & 71.3 & -- \\
\newcite{luong:2013}  & 34.36 & -- & 48.48  &  64.58 & --\\ 
\newcite{qiu-EtAl:2014} & 32.13 & -- & 53.40 & 65.19 & -- \\
\newcite{neelakantan:2014} & -- & -- & 65.5 &  69.2 & --\\
\newcite{chen-liu-sun:2014} & -- & -- & 64.2 & -- & --\\
\newcite{Hill:2015} & -- & \textbf{41.4} & -- & 65.5 & 69.9\\
\newcite{vilnis2015} & -- & 32.23 & -- & 65.49 & 71.31\\
\newcite{schnabel2015eval} & -- & -- & -- & 64.0 & 70.7 \\
\newcite{rastogi2015NAACL} & 32.9 & 36.7 & 65.6 & 70.8 & 73.9\\
\newcite{flekova-gurevych:2016:P16-1} & -- & -- & -- & -- & 74.26 \\
\hline
\hline
Word2Vec Skip-gram  & 32.64 & 38.20 &66.37 & {71.61} & 75.49 \\
\hline
$\textsc{mswe-1}_{50}$ & 34.85 & 38.77 & \textbf{66.83} & \textbf{72.40} & \underline{76.23} \\
$\textsc{mswe-1}_{200}$  & \underline{35.27} & 38.70 &  \underline{66.80} & 72.05 & 76.05 \\
\hline
$\textsc{mswe-2}_{50}$ & 34.98 & 38.79 & 66.61 & 71.71 & 75.90 \\
$\textsc{mswe-2}_{200}$  & \textbf{35.56}$^\star$ & \underline{39.19}$^\star$ & 66.65 & \underline{72.29} & \textbf{76.37}$^\star$ \\
\hline
\hline
\end{tabular}
}
\caption{Spearman's rank correlation ($\rho \times 100$) for the word similarity task when using $GlobalSim$. Subscripts 50 and 200 denote the online LDA topic model trained with $T=50$ and $T=200$ topics, respectively. 
$^\star$ denotes that our best score is significantly higher than the score of the baseline (with $p <$ 0.05, online toolkit from \url{http://www.philippsinger.info/?p=347}). Scores in \textbf{bold} and \underline{underline} are the best and second best scores.}
\label{tab:wordsimilarity}
\end{table}
}

\subsubsection{Results for word similarity}
\label{sec:wordsim}

Table \ref{tab:wordsimilarity} compares the evaluation results of \textsc{mswe} with results reported in prior work on the standard word similarity task when using $GlobalSim$.
We use subscripts 50 and 200 to denote the topic model trained with $T=50$ and $T=200$ topics, respectively.
Table \ref{tab:wordsimilarity} shows that our model  outperforms the baseline Word2Vec Skip-gram model (in fifth row from bottom).
Specifically, on the \textsc{rw} dataset, \textsc{mswe} obtains a significant improvement of $2.92$
in the Spearman's rank correlation (which is about 8.5\% relative improvement).

Compared to the published results, \textsc{mswe} obtains the highest accuracy on the \textsc{rw}, \textsc{scws}, \textsc{ws353} and \textsc{men} datasets, and achieves the second highest result on the \textsc{SimLex} dataset.
These indicate that \textsc{mswe} learns better representations for words taking into account different meanings.


{
\begin{table}[!t]
\centering
\setlength{\tabcolsep}{0.25em}
\resizebox{7.5cm}{!}{
\begin{tabular}{l|c|c}
\hline
Model & AvgSim & AvgSimC \\
\hline
\hline
\newcite{Huang:2012}  & 62.8  & 65.7   \\
\newcite{neelakantan:2014}   & \textbf{67.3}  & \textbf{69.3}  \\
\newcite{chen-liu-sun:2014}  & 66.2   & \underline{68.9}  \\
\newcite{chen:2015} & 65.7  & 66.4  \\
\newcite{Wu:2015} & -- & 66.4  \\
\newcite{jauhar:2015} & -- & 65.7  \\
\newcite{chengJianpeng:2015} & 62.5   & -- \\
\newcite{iacobacci:2015} & 62.4   & --\\
\newcite{Cheng:2015} & -- & 65.9 \\
\hline
\hline
$\textsc{mswe-1}_{50}$ & 66.6  & {66.7} \\
$\textsc{mswe-1}_{200}$ & \underline{66.7} & 66.6 \\
\hline
$\textsc{mswe-2}_{50}$ & 66.4  & 66.6 \\
$\textsc{mswe-2}_{200}$ & 66.6  & 66.6 \\
\hline
\hline
\end{tabular}
}
\caption{Spearman's rank correlation ($\rho \times 100$) on   \textsc{scws}, using $AvgSim$ and $AvgSimC$.}
\label{tab:scws}
\end{table}
}

\subsubsection{Results for contextual word similarity}
We evaluate our model \textsc{mswe} by using $AvgSim$ and $AvgSimC$ on the benchmark \textsc{scws} dataset which considers effects of the contextual information on the word similarity task.
As shown in Table \ref{tab:scws}, \textsc{mswe}  scores better than the closely related model proposed by \newcite{Cheng:2015} and generally obtains good results for this context sensitive dataset.
Although we produce better scores than \newcite{neelakantan:2014} and \newcite{chen-liu-sun:2014} when using $GlobalSim$, we are outperformed by them when using $AvgSim$ and $AvgSimC$. 
\newcite{neelakantan:2014} clustered the embeddings of the context words around each target word to predict its sense and \newcite{chen-liu-sun:2014} used pre-trained word embeddings to initialize vector representations of senses taken from WordNet, while we use a fixed number of topics as senses for words in \textsc{mswe}.


\subsection{Word Analogy}

{
\begin{table}[!t]
\centering
\begin{tabular}{l|c}
\hline
Model & Accuracy (\%)\\
\hline
\hline
\newcite{Pennington2014} & \textbf{70.3} \\
\newcite{baroni:2014} & 68.0 \\
\newcite{neelakantan:2014} & 64.0 \\
\newcite{Ghannay:2016} & 62.3 \\
\hline
\hline
Word2Vec Skip-gram  & 68.6 \\
\hline
\hline
$\textsc{mswe-1}_{50}$ & 69.6 \\
$\textsc{mswe-1}_{200}$ & \underline{69.9} \\
\hline
$\textsc{mswe-2}_{50}$ & 69.7 \\
$\textsc{mswe-2}_{200}$ & 69.5 \\
\hline
\hline
\end{tabular}
\caption{Accuracies  for the word analogy task. 
All our results are significantly higher than the result of Word2Vec Skip-gram (with two-tail $p < 0.001$ using  McNemar's test). 
\newcite{Pennington2014} used a larger training corpus of 1.6B words.
}
\label{tab:wordanalogy}
\end{table}
}

We evaluate the embedding models on the word analogy task introduced by \newcite{MikolovCoRR2013}. 
The task aims to answer questions in the form of ``$a$ is to $b$ as $c$ is to \_ $?$'', denoted as ``\emph{a : b $\rightarrow$ c : ?}'' (e.g., ``\emph{Hanoi : Vietnam  $\rightarrow$ Bern : ?}'').
There are 8,869 semantic and 10,675 syntactic questions grouped into 14 categories.
Each question is answered by finding the most suitable word closest to ``\emph{$\boldsymbol{v}_{b} - \boldsymbol{v}_{a} + \boldsymbol{v}_{c}$}'' measured by the cosine similarity.
The answer is correct only if the found closest word is exactly the same as the gold-standard (correct) one for the question. 

We report accuracies in Table \ref{tab:wordanalogy} and show that \textsc{mswe} achieves better results in comparison with the baseline Word2Vec Skip-gram. 
In particular, \textsc{mswe} reaches the accuracies of around 69.7$\%$ which is higher than the accuracy of 68.6$\%$ obtained by Word2Vec Skip-gram.

\section{Conclusions}
\label{sec:conclusion}

In this paper, we described a mixture model for learning multi-sense embeddings. 
Our model induces mixture weights to represent a word given context based on a mixture of its sense representations. 
The results show that our model scores better than Word2Vec, and produces highly competitive results on the standard evaluation tasks.  
In future work, we will explore better methods for taking into account the contextual information. 
We also plan to explore different approaches to compute the mixture weights in our model.
For example, if there is a large sense-annotated corpus available for training, the mixture weights could be defined based on the frequency (sense-count) distributions, instead of using the probability distributions produced by a topic model.
Furthermore, it is possible to consider the weights of senses as additional model parameters to be then learned during training.

\section*{Acknowledgments}
This research was funded by the German Research Foundation (DFG) as part of SFB 1102 ``Information Density and Linguistic Encoding''.
We would like to thank anonymous reviewers for their helpful comments.

\bibliographystyle{acl_natbib}
\bibliography{references}

\end{document}